\documentclass{article}

\usepackage{arxiv}
\usepackage[utf8]{inputenc}
\usepackage[T1]{fontenc}
\usepackage{hyperref}
\usepackage{url}
\usepackage{booktabs}
\usepackage{amsfonts}
\usepackage{amsmath}
\usepackage{amssymb}
\usepackage{microtype}
\usepackage{graphicx}
\usepackage{multirow}
\usepackage{float} 
\usepackage{natbib}
\usepackage{array}
\usepackage{xcolor}
\usepackage{pgfplots}
\usepgfplotslibrary{groupplots}
\pgfplotsset{compat=1.18}
\usepackage{tikz}
\usepackage[export]{adjustbox}
\usepackage{caption}
\usepackage[printonlyused]{acronym}

\title{Conditional Flow Matching for ML-Based Inverse Design Problems}

\author{%
  Juliana Felder$^{1}$\thanks{Equal contribution.} \quad
  \textbf{Milad Habibi}$^{2}$\footnotemark[1] \quad
  \textbf{Soheyl Massoudi}$^{1}$ \quad
  \textbf{Mark Fuge}$^{1}$\\
  $^{1}$ETH Zürich\\
  $^{2}$University of Maryland, College Park\\
  \texttt{feldej@ethz.ch} \quad \texttt{mafuge@ethz.ch}
  \vspace{-5mm}
}

\renewcommand{\shorttitle}{Flow Matching for Engineering Inverse Design}

\hypersetup{
  hidelinks,
pdftitle={Conditional Flow Matching for ML-Based Inverse Design Problems},
pdfauthor={Juliana Felder, Milad Habibi, Soheyl Massoudi, Mark Fuge},
pdfkeywords={inverse design, flow matching, engineering optimization, EngiBench, generative models},
}

\begin{document}
\newacro{CFM}{conditional flow matching}
\newacro{COG}{cumulative optimality gap}
\newacro{FOG}{final optimality gap}
\newacro{NFE}{number of function evaluations}
\newacro{cGAN}{conditional generative adversarial network}
\newacroplural{cGAN}[cGANs]{conditional generative adversarial networks}
\newacro{ICFM}[I-CFM]{independent conditional flow matching}
\newacro{GAN}{generative adversarial network}
\newacroplural{GAN}[GANs]{generative adversarial networks}
\newacro{MMD}{maximum mean discrepancy}
\newacro{DPP}{determinantal point process}
\newacro{ODE}{ordinary differential equation}
\newacroplural{ODE}[ODEs]{ordinary differential equations}
\newacro{RK4}{classical fourth-order Runge--Kutta method}
\newacro{DDPM}{denoising diffusion probabilistic model}
\newacro{PDE}{partial differential equation}
\newacroplural{PDE}[PDEs]{partial differential equations}
\maketitle

\begin{abstract}
Engineering inverse design is often limited by the high computational cost of iterative solvers for optimization problems constrained by \acp{PDE} and by their sensitivity to initialization. Deep generative models can produce candidate designs without rerunning the simulator at inference time. \Acp{GAN} sample in one forward pass, whereas diffusion models require iterative reverse-time integration.

In this work, we add \ac{CFM} to EngiOpt and compare it with a conditional diffusion model and a \ac{cGAN} on structural (\texttt{beams2d}) and thermal (\texttt{heatconduction2d}) benchmarks from EngiBench using the same downstream optimization protocol. We use \ac{COG} and \ac{FOG} as the primary metrics for evaluating the generated designs as warm starts for gradient-based refinement. On the evaluated EngiOpt implementations and two EngiBench tasks, \acs{CFM} achieves the lowest measured \acs{COG}, \acs{FOG}, \ac{MMD}, and volume-fraction deviation on both tasks. \acs{CFM} has mean volume-fraction deviations of $0.4\%$ and $1.0\%$ on \texttt{beams2d} and \texttt{heatconduction2d}, respectively, compared with $3.8\%$ and $11.2\%$ for diffusion. At Euler $s=16$, \acs{CFM} achieves $53.2$ samples/s on \texttt{beams2d} --- about $66\times$ the measured throughput of the evaluated diffusion baseline using 1000 network evaluations under the same timing protocol --- with \acs{COG} $1.182 \pm 3.126$, compared with $1.173 \pm 3.100$ for Euler $s=32$.

Across the two tasks, \acs{CFM} produces warm starts with lower measured \acs{COG} than both baselines and uses fewer network evaluations than diffusion.

\end{abstract}
\acresetall


\section{Introduction}
\label{sec:introduction}

Engineering inverse design seeks geometries that satisfy target physical performance criteria --- a task traditionally solved by iterative gradient-based optimization constrained by expensive physics simulations~\citep{peckham2025artificial, habibi2026when}. Each design update requires a simulator call, making it expensive to solve many related design instances~\citep{giannone2023aligning}. Generative models offer a data-driven alternative: trained on reference designs produced by numerical optimization, they approximate $p(x \mid c)$ and generate candidates at inference time without rerunning the simulator~\citep{regenwetter2022deep, maze2023diffusion}. The aim is not to generate a fully converged design, but to provide an initialization that the downstream optimizer can refine under the prescribed physical conditions.

Prior studies report improved sample quality for diffusion models over \acp{cGAN}~\citep{maze2023diffusion, zhang2025conditional, habibi2024inverse}, but hundreds of network evaluations per sample can make generation slow. We introduce \ac{CFM} as a new baseline in EngiOpt~\citep{felten2025engibench} --- alongside existing \ac{cGAN} and diffusion implementations --- replacing stochastic denoising with deterministic velocity regression~\citep{lipman2023flowmatching}. Because \ac{CFM} and diffusion share the same conditional U-Net backbone, the comparison limits differences in model architecture and focuses on the training objective and sampling procedure.

We ask whether \ac{CFM} can generate useful warm starts with fewer network evaluations than diffusion and lower seed-to-seed variability than the evaluated \ac{cGAN} baseline. Our contributions are:
\begin{enumerate}
    \item A \ac{CFM} baseline for EngiOpt/EngiBench, evaluated against
          conditional diffusion and \ac{cGAN} using the same data splits,
          checkpoint-selection procedure, and downstream optimizer.
    \item An evaluation protocol centered on \ac{COG} and \ac{FOG},
          measuring downstream warm-start utility via optimization
          trajectories rather than distributional metrics alone.
    \item A solver and \ac{NFE} ablation showing \texttt{beams2d} \ac{COG}
          of $1.182 \pm 3.126$ at \ac{NFE}$=16$ and $1.173 \pm 3.100$
          at \ac{NFE}$=32$, while \ac{NFE}$=16$ reaches approximately
          $64\times$--$66\times$ the measured diffusion throughput.
\end{enumerate}

\section{Related Work}
\label{sec:related}

Physics-based topology optimization repeatedly solves governing equations while updating a material distribution. These repeated analyses make solving many related design instances expensive~\citep{sigmund2013topology}. This cost has motivated machine-learning methods that predict or generate candidate designs~\citep{regenwetter2022deep}. We review learned warm starts, generative inverse-design models, diffusion models, and flow matching.

\paragraph{Inverse Design}
Inverse design seeks a geometry or material distribution that satisfies prescribed physical requirements. Applications include structural and thermal topology optimization, nanophotonics, and aerodynamic design~\citep{felten2025engibench,molesky2018nanophotonics,shirvani2023machine}. Structural and thermal topology optimization are commonly formulated as \acp{PDE}-constrained problems. Density-based methods such as SIMP solve them iteratively by alternating physical analysis and material updates~\citep{sigmund200199}. Although effective, this procedure can be expensive when designs are required for many conditions.

\paragraph{Learned Warm Starts}
Machine-learning models can reduce this cost by mapping design conditions to candidate material distributions. Direct prediction has been studied for structural and thermal topology optimization~\citep{li2019noniterative,lin2025intelligent}. Other methods use the generated design as an initialization and leave final physical refinement to a conventional optimizer~\citep{giannone2023aligning,habibi2026when}. Visual similarity to a reference design does not necessarily imply comparable physical performance~\citep{habibi2025mean}. We therefore assess learned initializations through their effect on downstream optimization, alongside distributional metrics.

\paragraph{Generative Models for Engineering Design}
Engineering inverse-design problems can admit several designs that satisfy the same requirements. Deep generative models provide a way to represent this conditional design distribution. \Acp{cGAN} have been applied to topology optimization using loads, boundary conditions, or physical fields as inputs~\citep{mirza2014cgan,nie2021topologygan,hertlein2021amgan}. They generate designs in one forward pass, but adversarial training can be unstable and susceptible to mode collapse~\citep{goodfellow2014gan}. Variational and other generative models have also been used to generate multiple candidates for subsequent evaluation or refinement~\citep{oh2019deep,yamasaki2021datadriven}.

\paragraph{Diffusion Models}
Diffusion models avoid adversarial training by learning to reverse a gradual noising process~\citep{ho2020ddpm}. TopoDiff reports up to an eightfold reduction in physical-performance error and elevenfold fewer non-manufacturable designs than the adversarial baselines considered in that study~\citep{maze2023diffusion}. Related diffusion approaches have been applied to lattice and structural-component generation~\citep{zhang2025conditional,herron2024latent}. The relative performance of diffusion and adversarial models nevertheless depends on the task and available training data~\citep{habibi2024inverse}. Diffusion sampling also requires repeated network evaluations, motivating comparisons that consider both design quality and generation cost.

\paragraph{Flow Matching}
Flow matching trains continuous normalizing flows without simulating complete trajectories during training~\citep{lipman2023flowmatching}. A continuous normalizing flow transports a sample $x_0$ from a source distribution $p_0$ to a sample $x_1$ from the data distribution through a time-dependent vector field $v_t$:
\begin{equation}
\frac{\mathrm{d}x_t}{\mathrm{d}t}=v_t(x_t), \qquad x_0\sim p_0.
\end{equation}
Classical continuous normalizing flows generally require solving and differentiating through this \ac{ODE} during training. Flow matching instead regresses the vector field associated with a prescribed probability path~\citep{albergo2022building}.

Conditional flow matching constructs paths whose target velocities can be computed from sampled source and target endpoints. In \ac{ICFM}, these endpoints are sampled independently. Optimal-transport couplings can instead reduce path curvature and objective variance~\citep{tong2023improving}. The linear path, target velocity, and training objective used in this work are defined in Section~\ref{sec:method}. Related formulations include stochastic interpolants and rectified flow~\citep{albergo2025stochasticinterpolants,liu2023rectifiedflow}.

Recent work applies flow matching to airfoil and wing inverse design~\citep{yang2026physicsguided,zhao2026flowmatching}, ship-propeller design~\citep{kruger2026generative}, fluid-field generation~\citep{kashefi2026flow}, and topology optimization~\citep{xiao2026trajectory,rashed2026generalization}. \citet{kruger2026well} compare generative models on engineering inverse problems, while \citet{de2026generative} study a flow-matching method for high-dimensional inverse design with abstention.

\paragraph{Scope of This Study}
Previous work shows that flow matching can be used for engineering design, including topology optimization. We compare the linear-path \ac{ICFM} implementation described in Section~\ref{sec:method} with the existing diffusion and \ac{cGAN} implementations in EngiOpt. All methods use the same EngiBench data splits, matched condition--reference-design pairs, validation-only checkpoint selection, downstream optimizer, and metric definitions. \ac{CFM} and Diffusion additionally share the same conditional U-Net architecture. We evaluate structural and thermal tasks and report \ac{COG} and \ac{FOG} computed from the downstream optimization trajectories together with measured sampling cost.

\section{Method}
\label{sec:method}

\paragraph{Problem Formulation}
Each problem instance is defined by a condition vector $c \in \mathbb{R}^{n_c}$ encoding problem-specific boundary conditions, and a target 2D density field $x \in [0,1]^{H \times W}$, where each entry represents local material density (0 = void, 1 = solid). The dataset $\mathcal{D} = \{(c^{(i)}, x_{\mathrm{ref}}^{(i)})\}_{i=1}^{N}$ consists of $N$ pairs of boundary conditions and their corresponding reference designs. Our objective is to learn a conditional generative model that approximates the distribution $p(x \mid c)$, whose samples serve as initial designs for downstream physics-based optimization.

\paragraph{Engineering Design Problems and Data Representation}
EngiBench provides common Python interfaces for loading benchmark datasets and running their simulators and optimizers~\citep{felten2025engibench}. We evaluate structural compliance optimization on \texttt{Beams2D} and thermal compliance optimization on \texttt{HeatConduction2D}. Full problem documentation is available in the EngiBench problem registry.\footnote{\url{https://engibench.ethz.ch/}}

\begin{itemize}
\item \textbf{\texttt{Beams2D}:} Structural compliance minimization for the right-half MBB beam on a $100 \times 50$ grid ($x \in [0,1]^{100 \times 50}$, $c \in \mathbb{R}^4$). The dataset provides 4,851 optimized reference designs: 3,880 training, 728 validation, and 243 test designs. Condition variables $c$ include volume fraction (\textit{volfrac}), minimum feature thickness (\textit{rmin}), fractional force location (\textit{forcedist}), and an overhang manufacturability flag.
\item \textbf{\texttt{HeatConduction2D}:} Thermal compliance minimization under material-budget and boundary-condition constraints on a $101 \times 101$ grid ($x \in [0,1]^{101 \times 101}$, $c \in \mathbb{R}^2$). The dataset contains 441 optimized reference designs: 361 training, 40 validation, and 40 test designs. Condition variables $c$ include the volume limit on material distribution and the length of the adiabatic region on the bottom side of the domain.
\end{itemize}

\paragraph{Preprocessing and Postprocessing Pipelines}

Preprocessing is model-specific. For \ac{CFM}, the design fields are min-max normalized using statistics computed on the training split, $x_{\text{norm}} = (x - x_{\min}) / (x_{\max} - x_{\min})$, while the condition vectors are left unchanged. Since the training data is already in $[0,1]$, this normalization is approximately the identity; generated outputs are clipped directly to $[10^{-3}, 1]$ before evaluation and use as optimizer warm starts. Diffusion designs are instead rescaled to $[-1,1]$ for training, matching the standard \ac{DDPM} convention~\citep{ho2020ddpm}, and generated outputs are rescaled back to the physical density range before being clipped to $[10^{-3},1]$. For \ac{cGAN}, the generator uses a sigmoid output, so generated designs are already in $[0,1]$ before applying the same clipping step.

\subsection{\acs{CFM} Architecture}
We use a 2D conditional U-Net as the backbone for \ac{ICFM}. The U-Net takes a noisy design $x_t \in \mathbb{R}^{1 \times H \times W}$ and condition $c$ as input, and outputs a velocity field $v_\theta(x_t, t, c) \in \mathbb{R}^{1 \times H \times W}$. Concretely, we implement the \texttt{UNet2DConditionModel} from the Hugging Face Diffusers library~\citep{von-platen-etal-2022-diffusers}. It has four encoder-decoder stages at channel depths $(32, 64, 128, 256)$, with conditions injected via cross-attention at every block.

As discussed in Section~\ref{sec:related}, we use \ac{ICFM} where Gaussian noise $x_0 \sim \mathcal{N}(0,I)$ and target designs $x_1$ are paired independently at random. For a continuous time variable $t \sim\mathcal{U}(0, 1)$, we define the linear interpolation path $x_t = (1 - t)x_0 + t x_1$ with a target velocity $v^\star = x_1 - x_0$. The model predicts $v_\theta(x_t, t, c)$ and is trained with a mean-squared error objective:
\begin{equation}
\mathcal{L}_{\mathrm{FM}} = \mathbb{E}_{x_0,x_1,t}\left[\|v_\theta(x_t,t,c)-(x_1-x_0)\|_2^2\right]
\end{equation}

\paragraph{Inference and Solvers}
During inference, designs are generated by integrating the learned velocity field from $t=0$ to $t=1$ using a fixed-step \ac{ODE} solver. To evaluate the trade-off between sampling speed and solution quality, we benchmark three solvers: first-order Euler, second-order Midpoint, and the \ac{RK4}. We compare them at \ac{NFE} budgets of 16, 32, and 48. This corresponds to step counts of $\{16, 32, 48\}$ for Euler, $\{8, 16, 24\}$ for Midpoint, and $\{4, 8, 12\}$ for \ac{RK4}.

\subsection{Baselines}
To reduce architectural differences between \ac{CFM} and Diffusion, both use the same \texttt{UNet2DConditionModel} backbone and channel schedule $(32, 64, 128, 256)$, with the same number of trainable parameters. Diffusion is trained with a linear noise schedule ($\beta_1=10^{-4}$, $\beta_T=0.02$, $T=1000$ steps) to predict noise $\epsilon$ and is sampled via \ac{DDPM} ancestral sampling over 1000 reverse steps. The \ac{cGAN} uses the existing EngiOpt adversarial architecture. The generator takes noise $z \sim \mathcal{N}(0,I)$ (latent dimension $z_{\mathrm{dim}}=32$) and condition $c$ expanded to a spatial map; both are processed by parallel \texttt{ConvTranspose2d} stems before concatenation into $F_0=256$ channels. Four upsampling stages progress spatially from $7\times7$ to the target resolution, with BatchNorm and ReLU activations, and a final sigmoid layer producing $\hat{x} \in [0,1]^{1 \times H \times W}$ before clipping to $[10^{-3}, 1]$ for evaluation. The discriminator mirrors this conditioning strategy with parallel image and condition stems downsampled to a single real-or-generated prediction.

Detailed optimizer, hardware, and training settings are provided in Appendix~\ref{app:training-configuration}.

\subsection{Experimental Protocol and Evaluation Metrics}
Rather than treating the generative model as a stand-alone design generator, we use it as a warm-start prior for gradient-based topology optimization. The benchmark tasks, data splits, conditions, downstream optimizer, and evaluation metrics are fixed across methods. \acs{CFM} and Diffusion also share the same U-Net architecture, whereas \ac{cGAN} uses its existing EngiOpt architecture.

In this pipeline, the generative model receives problem-specific conditions $c^{(i)}$ and produces an initial candidate design $x_0^{(i)}$ for each test condition. This warm start is passed to the problem-specific benchmark optimizer provided by EngiBench~\citep{felten2025engibench}, which refines the design under the same condition $c^{(i)}$, producing an optimization trajectory $\{x_t^{(i)}\}_{t=0}^{T}$. The trajectory is then evaluated against the paired dataset reference design $x_{\mathrm{ref}}^{(i)}$ using the metrics defined below. All methods use the same dataset splits and training seeds 1--10.

\paragraph{Training and Model Selection}
All models are trained for at most 500 epochs with a batch size of 32. For each training seed $s$, we sample 50 condition--reference-design pairs with replacement from the validation split using seed $s+123$; this set remains fixed throughout training. Every 10 epochs, generated designs for these conditions are scored by validation \ac{MMD}. Checkpoints become eligible for selection after epoch 80, and training stops after 25 consecutive validation checks without a lower \ac{MMD}.

Model selection uses only the validation split. We first retain the five eligible checkpoints with the lowest validation \ac{MMD}. Each checkpoint is then used to initialize the downstream optimizer on the same 50 validation pairs and with the same optimizer budget used for final evaluation. The checkpoint with the lowest validation \ac{COG} is selected. It is evaluated on 50 matched condition--reference-design pairs sampled with replacement from the test split using seed $s$, while generation uses seed $s+2000$. For a given problem and training seed, \ac{CFM}, Diffusion, and \ac{cGAN} therefore use the same validation and test pairs, reference designs, downstream optimizer, and metric definitions. Test results are not used for early stopping or checkpoint selection.

To measure how the generated warm start affects the downstream optimization trajectory under a fixed optimizer budget, let $\{x_t^{(i)}\}_{t=0}^{T}$ denote the trajectory starting from generated design $x_0^{(i)}$ under condition $c^{(i)}$, and let $f_{\mathrm{ref}}^{(i)} = f(x_{\mathrm{ref}}^{(i)}, c^{(i)})$ be the objective of the paired dataset design evaluated under the same condition. Note that $x_{\mathrm{ref}}^{(i)}$ is not a global optimum but a reference design produced by the benchmark optimizer.
\begin{itemize}
\item \textbf{\ac{COG}:} Measures the cumulative objective gap relative to the reference over the optimization trajectory:
\begin{equation}
\mathrm{COG}
=
\operatorname{median}_{i}
\sum_{t=1}^{T}
\left[
f(x_t^{(i)}, c^{(i)}) - f(x_{\mathrm{ref}}^{(i)}, c^{(i)})
\right].
\label{eq:cog}
\end{equation}
Intuitively, \ac{COG} accumulates the objective difference between the warm-started optimization trajectory and the paired reference design. A trajectory whose objective values remain close to the reference produces a small \ac{COG}. A negative \ac{COG} indicates that the cumulative objective difference over the trajectory is below zero relative to the dataset reference. Median aggregation is used for \ac{COG} because the cumulative trajectory sum is sensitive to outlier optimization runs.

\item \textbf{\ac{FOG}:} Measures the quality of the final design $x_T^{(i)}$ relative to the dataset reference:
\begin{equation}
  \mathrm{FOG}
=
\mathbb{E}_{i}
\left[
f(x_T^{(i)}, c^{(i)}) - f(x_{\mathrm{ref}}^{(i)}, c^{(i)})
\right].
  \label{eq:fog}
\end{equation}
\end{itemize}

Unlike \ac{COG}, \ac{FOG} considers only the final step of the optimization trajectory. A positive \ac{FOG} means the optimizer terminated above the reference objective. A negative \ac{FOG} means the warm-started optimizer found a solution better than the dataset reference. We report mean \ac{FOG} to summarize the average final objective difference across evaluated samples. For both \ac{COG} and \ac{FOG}, lower is better.

\paragraph{Distributional and Feasibility Metrics}
\ac{MMD} measures distributional agreement between generated and reference designs using a Gaussian kernel with problem-specific bandwidth $\sigma = 1.0$ for \texttt{Beams2D} and $10.0$ for \texttt{HeatConduction2D}. Because the bandwidth is selected separately for each design space, \ac{MMD} and \ac{DPP} values are comparable across methods only within the same problem, not across problems.

\ac{DPP} diversity measures sample diversity via the determinant of the kernel similarity matrix of the generated batch. Higher values indicate greater diversity. Constraint violation (Viol.) quantifies the deviation from the required material budget, calculated as the mean absolute error between the generated design's material fraction and the target volume fraction $v^{*(i)}$ specified in $c$:
\begin{equation}\mathrm{Viol.} = \mathbb{E}_i\bigl[|\bar{x}_0^{(i)} - v^{*(i)}|\bigr],
\end{equation}
where $\bar{x}_0^{(i)}$ denotes the mean density of the generated design. We also inspect the 2D density fields for visible discontinuities, fragmented material regions, and differences from the paired reference topology.

\section{Results}
\label{sec:results}

Results are reported over 10 random seeds. \ac{COG} is summarized as median $\pm$ standard deviation, consistent with Eq.~\ref{eq:cog}; all other metrics are summarized as mean $\pm$ standard deviation. We report \ac{CFM} results using the Euler solver with $s=32$ steps (\ac{NFE}$=32$) for the main comparison against Diffusion (\ac{NFE}$=1000$) and \ac{cGAN}. Detailed solver and step-count results are provided in Appendix~\ref{app:ablation}.

\begin{table}[H]
\centering
\small
\resizebox{\textwidth}{!}{\begin{tabular}{lcccc}
\toprule
Method & \multicolumn{2}{c}{\texttt{Beams2D}} & \multicolumn{2}{c}{\texttt{HeatConduction2D}} \\
\cmidrule(lr){2-3} \cmidrule(lr){4-5}
 & \acs{COG} $\downarrow$ & \acs{FOG} $\downarrow$ & \acs{COG} $\downarrow$ & \acs{FOG} $\downarrow$ \\
\midrule
\acs{CFM} (Euler $s=32$) & $\mathbf{1.173 \pm 3.100}$ & $\mathbf{-1.647 \pm 0.411}$ & $\mathbf{8.88\times10^{-5} \pm 6.84\times10^{-5}}$ & $\mathbf{1.32\times10^{-6} \pm 8.87\times10^{-7}}$ \\
Diffusion & $1.603 \pm 1.794$ & $-1.637 \pm 0.505$ & $5.81\times10^{-4} \pm 1.44\times10^{-4}$ & $1.54\times10^{-5} \pm 5.06\times10^{-6}$ \\
\acs{cGAN} & $22.493 \pm 819.282$ & $4.415 \pm 14.602$ & $5.50\times10^{-4} \pm 2.01\times10^{-4}$ & $1.21\times10^{-5} \pm 3.58\times10^{-6}$ \\
\bottomrule
\end{tabular}
}
\caption{Primary optimization-utility results. Lower is better for \acs{COG} and \acs{FOG}; the best value within each problem and metric is boldfaced.}
\label{tab:primary-optimization-results}
\end{table}

Table~\ref{tab:primary-optimization-results} shows that \ac{CFM} obtains the lowest measured \ac{COG} and \ac{FOG} on both benchmarks.
On \texttt{beams2d}, \ac{CFM} and Diffusion reach mean \ac{FOG} values of $-1.647$ and $-1.637$, respectively, indicating that their warm-started downstream optimizations outperform the paired dataset references on average.
\ac{CFM} achieves a lower \ac{COG} ($1.173 \pm 3.100$) than Diffusion ($1.603 \pm 1.794$) and a slightly lower \ac{FOG} ($-1.647 \pm 0.411$ versus $-1.637 \pm 0.505$).
The \ac{cGAN} baseline shows high seed-dependent variability on this task, as evidenced by its \ac{COG} of $22.493 \pm 819.282$ and mean \ac{FOG} of $4.415 \pm 14.602$.

On \texttt{heatconduction2d}, \ac{CFM} has the lowest values for both primary metrics; Diffusion's \ac{COG} is 6.54 times \ac{CFM}'s. \ac{cGAN} has a \ac{COG} of $5.50\times10^{-4} \pm 2.01\times10^{-4}$, compared with Diffusion's $5.81\times10^{-4} \pm 1.44\times10^{-4}$. \ac{CFM} has the lowest measured \ac{COG} on both tasks.

\begin{figure}[htbp]
  \centering
  \small
  \setlength{\tabcolsep}{3pt}
    \begin{tabular}{lcccc}
      & \textbf{Reference}
      & \textbf{\acs{CFM} (Euler $s=32$)}
      & \textbf{Diffusion}
      & \textbf{\acs{cGAN}} \\[4pt]

      \rotatebox[origin=c]{90}{\texttt{Beams2D}} &
      \includegraphics[height=2.2cm,valign=m]{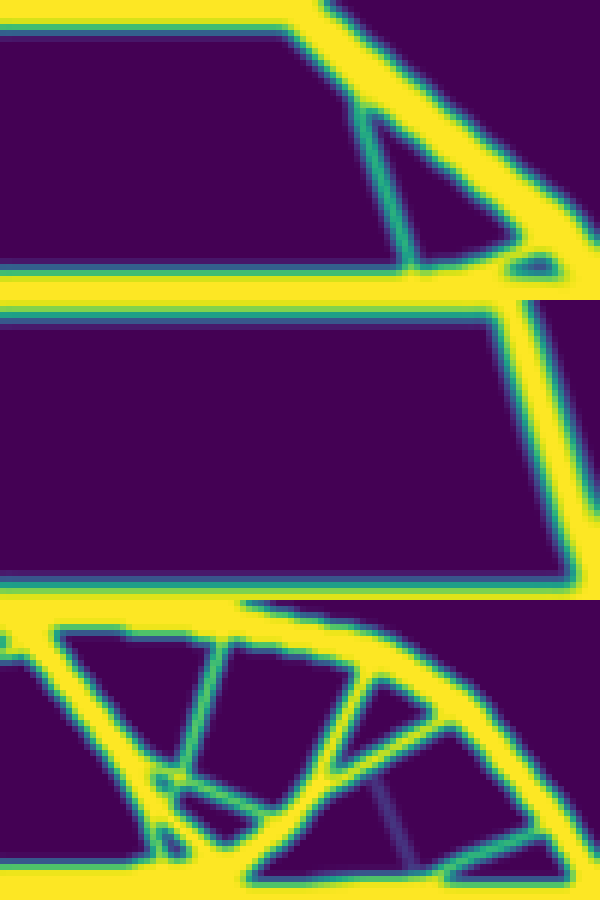} &
      \includegraphics[height=2.2cm,valign=m]{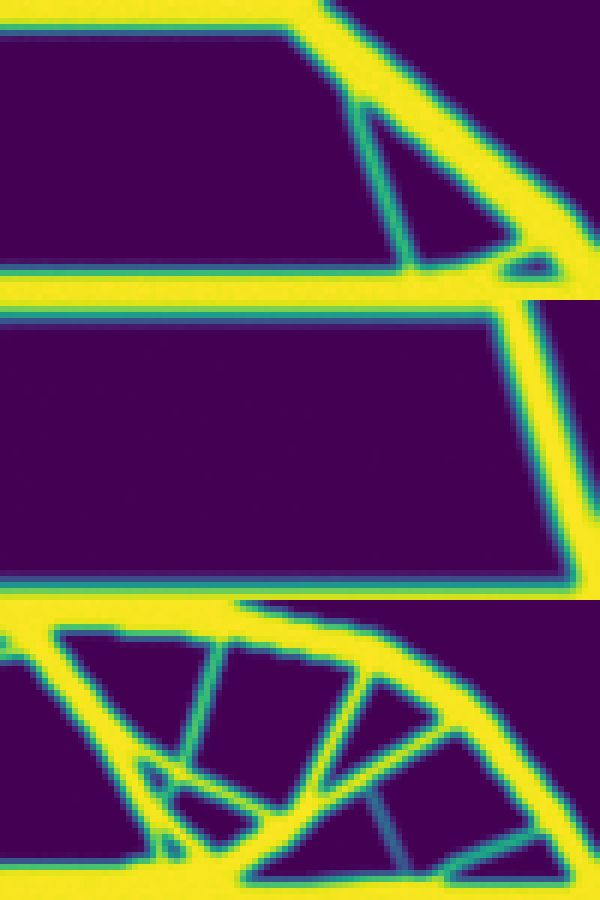} &
      \includegraphics[height=2.2cm,valign=m]{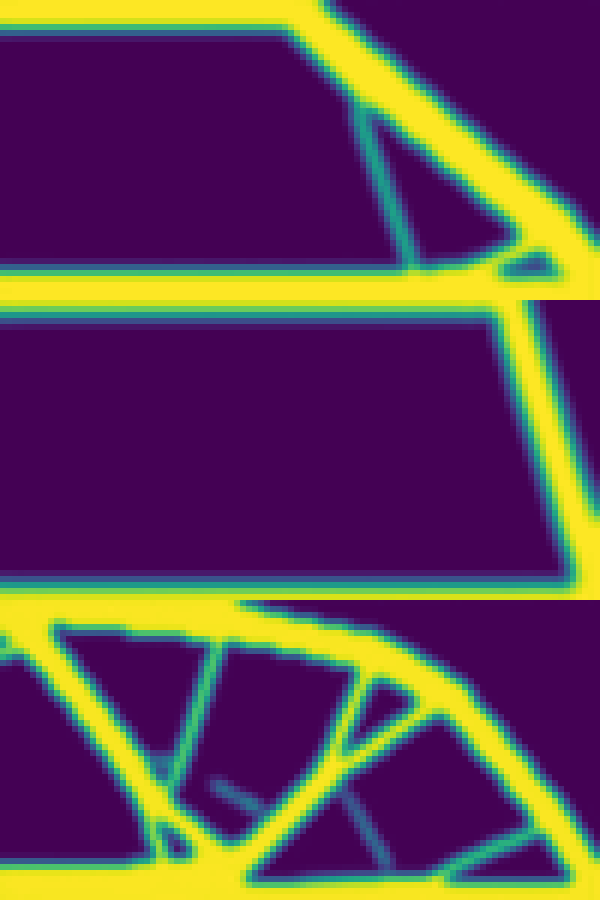} &
      \includegraphics[height=2.2cm,valign=m]{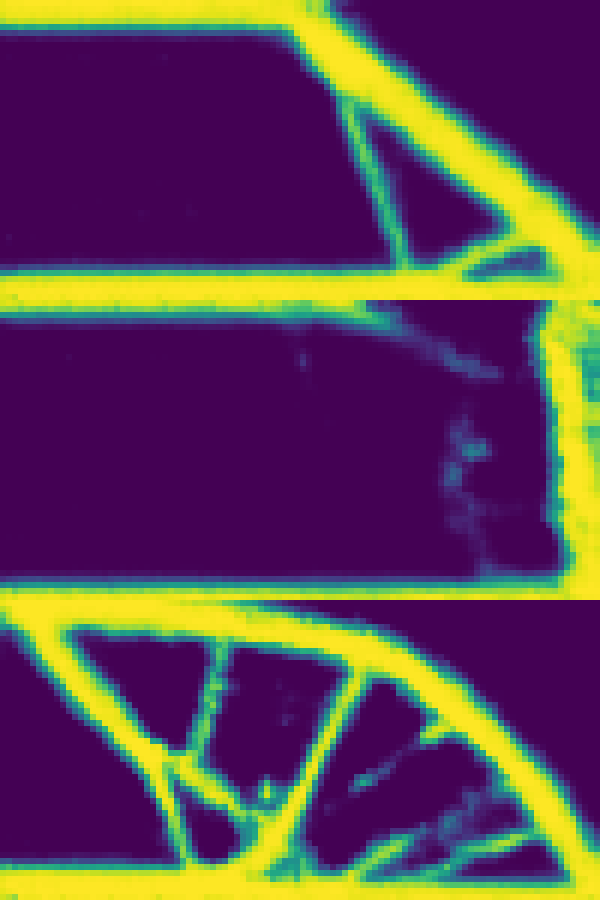} \\[8pt]

      \rotatebox[origin=c]{90}{\texttt{HeatCond2D}} &
      \includegraphics[height=3.2cm,valign=m]{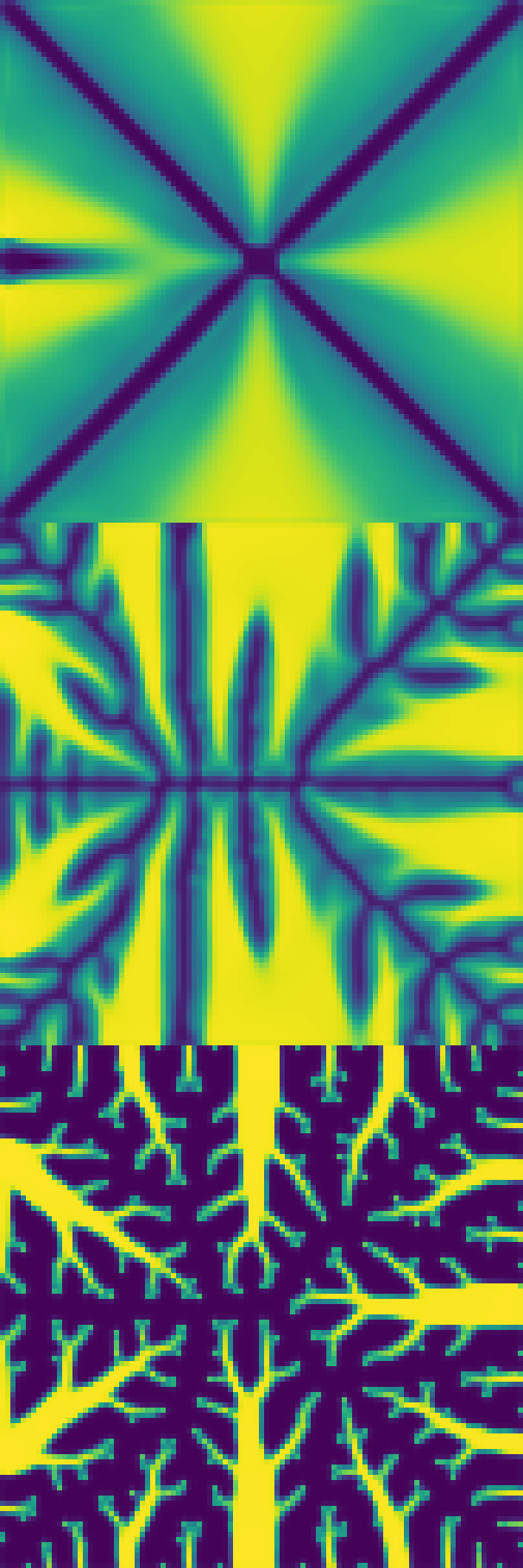} &
      \includegraphics[height=3.2cm,valign=m]{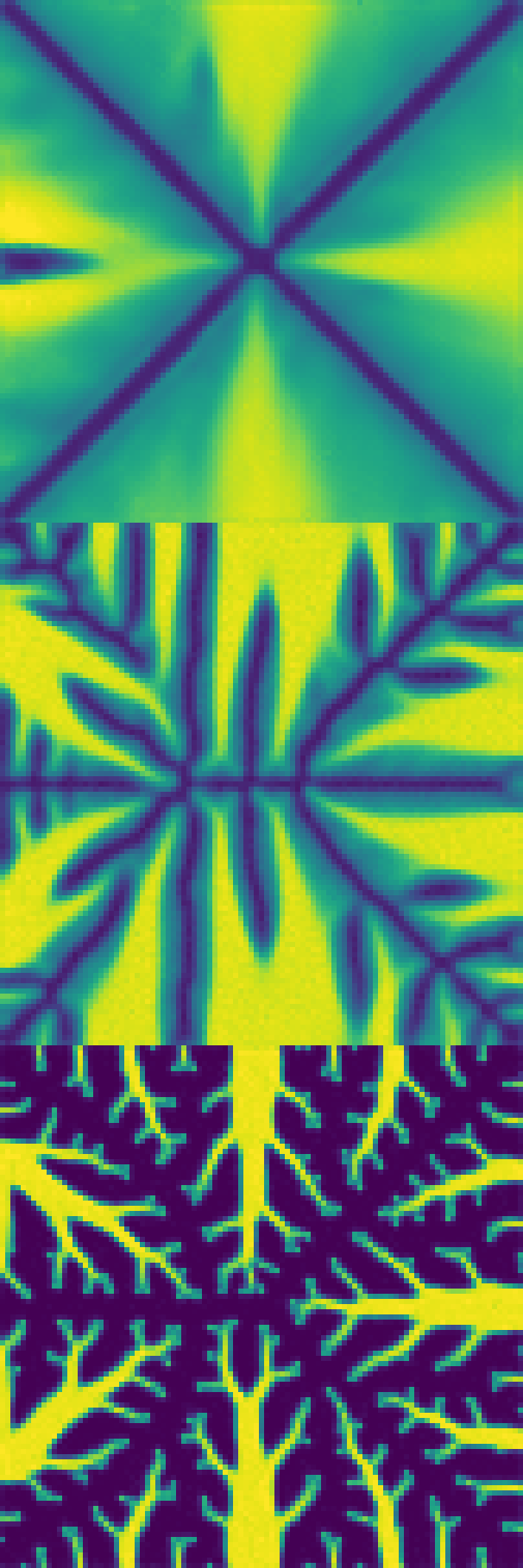} &
      \includegraphics[height=3.2cm,valign=m]{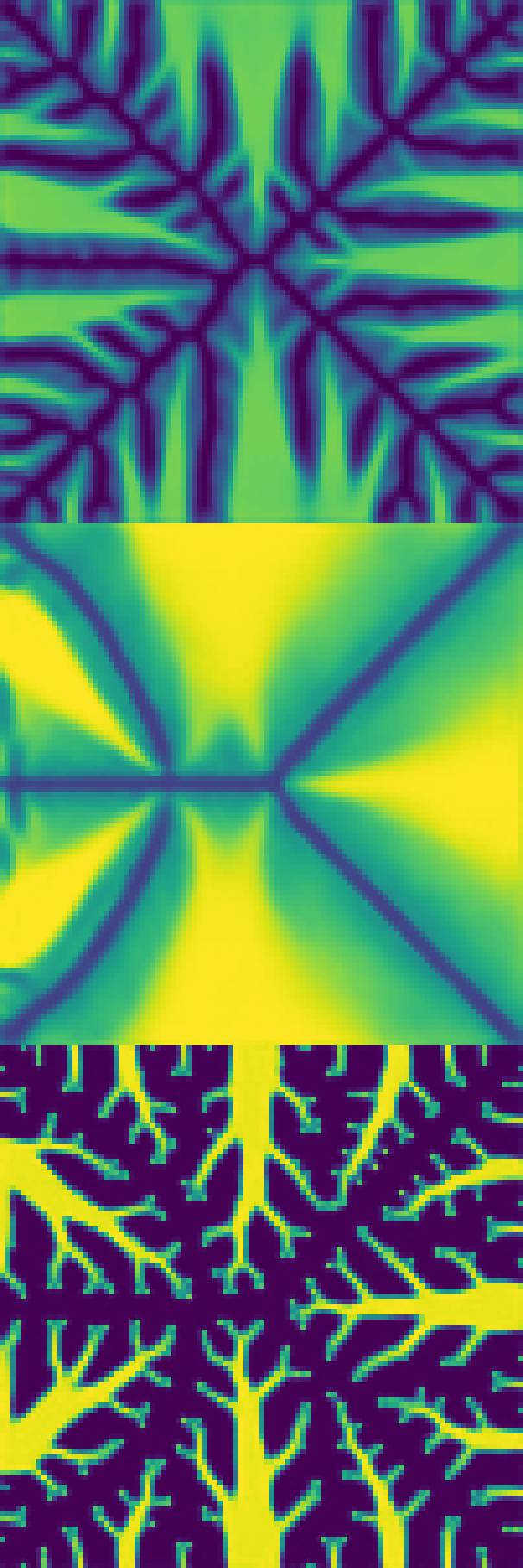} &
      \includegraphics[height=3.2cm,valign=m]{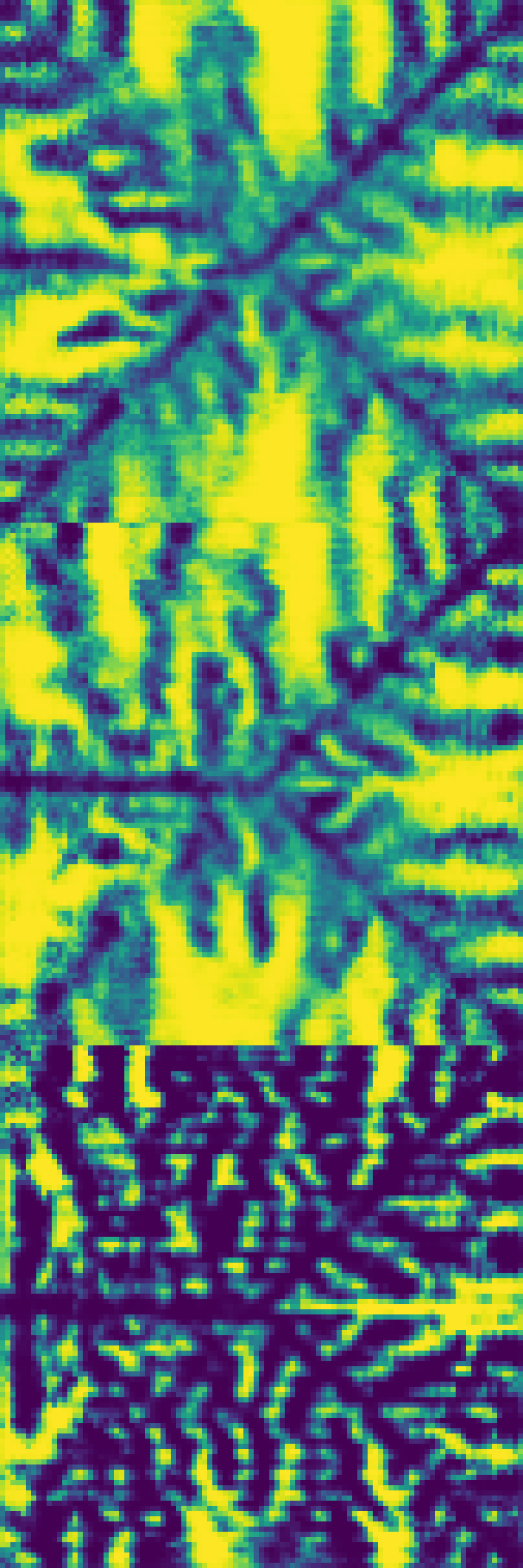} \\
    \end{tabular}
  \caption{Qualitative comparison under matched conditions. Each row corresponds to a benchmark problem; columns show the reference design and the warm start generated by each method. For \texttt{Beams2D} we visualize Seed 8, and for \texttt{HeatConduction2D} Seed 5; each row uses the same condition and seed across all displayed methods.}
  \label{fig:qualitative-results}
\end{figure}

\begin{table}[H]
\centering
\small
\resizebox{\textwidth}{!}{\begin{tabular}{lccc}
\toprule
Method & \acs{MMD} $\downarrow$ & \acs{DPP} $\uparrow$ & Viol. $\downarrow$ \\
\midrule
\multicolumn{4}{l}{\textit{\texttt{Beams2D}}} \\
\midrule
\acs{CFM} (Euler $s=32$)
  & $\mathbf{2.86\times10^{-2} \pm 2.06\times10^{-3}}$
  & $4.34\times10^{-1} \pm 2.69\times10^{-1}$
  & $\mathbf{3.60\times10^{-3} \pm 6.16\times10^{-4}}$ \\
Diffusion
  & $3.43\times10^{-2} \pm 4.19\times10^{-3}$
  & $\mathbf{6.49\times10^{-1} \pm 2.92\times10^{-1}}$
  & $3.82\times10^{-2} \pm 4.75\times10^{-2}$ \\
\acs{cGAN}
  & $4.50\times10^{-2} \pm 2.29\times10^{-3}$
  & $3.03\times10^{-1} \pm 4.40\times10^{-1}$
  & $1.62\times10^{-2} \pm 5.93\times10^{-3}$ \\
\midrule
\multicolumn{4}{l}{\textit{\texttt{HeatConduction2D}}} \\
\midrule
\acs{CFM} (Euler $s=32$)
  & $\mathbf{5.75\times10^{-2} \pm 8.35\times10^{-3}}$
  & $8.99\times10^{-3} \pm 1.35\times10^{-2}$
  & $\mathbf{9.96\times10^{-3} \pm 2.09\times10^{-3}}$ \\
Diffusion
  & $6.85\times10^{-2} \pm 6.52\times10^{-3}$
  & $\mathbf{2.91\times10^{-1} \pm 3.54\times10^{-1}}$
  & $1.12\times10^{-1} \pm 1.38\times10^{-2}$ \\
\acs{cGAN}
  & $1.20\times10^{-1} \pm 3.70\times10^{-2}$
  & $6.28\times10^{-2} \pm 9.81\times10^{-2}$
  & $1.53\times10^{-2} \pm 5.50\times10^{-3}$ \\
\bottomrule
\end{tabular}
}
\caption{Distributional and feasibility metrics. Lower is better for \acs{MMD} and Viol.; higher is better for \acs{DPP}. \acs{MMD} and \acs{DPP} are comparable across methods only within the same problem.}
\label{tab:secondary-results}
\end{table}

In the samples displayed in Figure~\ref{fig:qualitative-results}, \ac{CFM} retains the continuous members and sharp solid--void boundaries visible in the reference designs. Diffusion also produces sharp, high-contrast outputs, but its displayed \texttt{heatconduction2d} samples differ from the paired reference material distributions; this observation accompanies its larger average volume-fraction deviation in Table~\ref{tab:secondary-results}. In the displayed \texttt{beams2d} sample, \ac{cGAN} introduces a structural discontinuity absent from the reference. Its displayed \texttt{heatconduction2d} samples contain fragmented and isolated material regions.

Table~\ref{tab:secondary-results} details the secondary performance metrics. \ac{CFM} obtains the lowest \ac{MMD} and volume-fraction violation on both tasks, with volume-fraction violations of $\mathbf{3.60\times10^{-3}}$ for \texttt{beams2d} and $\mathbf{9.96\times10^{-3}}$ for \texttt{heatconduction2d}. Diffusion generates the highest \ac{DPP} values on both problems ($\mathbf{6.49\times10^{-1}}$ and $\mathbf{2.91\times10^{-1}}$), while also producing the highest volume-fraction deviations ($3.82\times10^{-2}$ and $1.12\times10^{-1}$). \ac{cGAN} reports lower volume-fraction violations than Diffusion ($1.62\times10^{-2}$ and $1.53\times10^{-2}$) alongside lower \ac{DPP} scores. On both tasks, the method with the highest \ac{DPP} does not achieve the lowest \ac{COG}.

\begin{table}[H]
\centering
\small
\resizebox{\linewidth}{!}{\begin{tabular}{llcc}
\toprule
Method & Problem & gen\_runtime\_sec $\downarrow$ & samples/s $\uparrow$ \\
\midrule
\multirow{2}{*}{\acs{CFM} (Euler $s=32$)}
  & \texttt{beams2d} & $1.876 \pm 0.022$ & $26.659 \pm 0.312$ \\
  & \texttt{heatconduction2d} & $4.092 \pm 0.052$ & $12.221 \pm 0.157$ \\
\midrule
\multirow{2}{*}{Diffusion}
  & \texttt{beams2d} & $61.903 \pm 1.027$ & $0.808 \pm 0.013$ \\
  & \texttt{heatconduction2d} & $131.74 \pm 1.53$ & $0.380 \pm 0.004$ \\
\midrule
\multirow{2}{*}{\acs{cGAN}}
  & \texttt{beams2d} & $4.382\times10^{-4} \pm 5.416\times10^{-6}$ & $114{,}107 \pm 1{,}419$ \\
  & \texttt{heatconduction2d} & $4.419\times10^{-4} \pm 5.404\times10^{-6}$ & $113{,}170 \pm 1{,}398$ \\
\bottomrule
\end{tabular}
}
\caption{Sampling throughput, measured as generated samples per second for the model generation call only. Higher is better.}
\label{tab:efficiency-results}
\end{table}

Table~\ref{tab:efficiency-results} reports generation throughput, defined as $n_\text{samples}/t_\text{gen}$. Generation time was measured separately from data loading, clipping, metric computation, optimizer refinement, and logging. For each method, we timed only the model sampling call on batches of $n_\text{samples}=50$ test conditions on a single NVIDIA RTX 4090 GPU, using CUDA synchronization before and after the timed region to avoid asynchronous-kernel timing artifacts. We used one warm-up call followed by three timed repeats for \ac{CFM} and Diffusion. Because the \ac{cGAN} forward pass is sub-millisecond, we used 1000 warm-up calls and 1000 timed repeats. For each seed, we take the median synchronized runtime across repeats and report the mean $\pm$ standard deviation over 10 seeds. \ac{cGAN} is fastest because it requires one forward pass, but it does not obtain the lowest \ac{COG} or \ac{MMD} on either task. \ac{CFM} at Euler $s=32$ is approximately $33.0\times$ faster than Diffusion on \texttt{beams2d} ($1.876$ s versus $61.903$ s) and $32.2\times$ faster on \texttt{heatconduction2d} ($4.092$ s versus $131.74$ s), while obtaining the lowest measured \ac{COG} on both tasks.

The solver-step ablation in Appendix~\ref{app:ablation} shows that, on \texttt{beams2d}, Euler and Midpoint obtain \ac{COG} values ranging from $1.134$ to $1.182$ across the evaluated \ac{NFE} budgets. At \ac{NFE}$=16$, \ac{RK4} ($s=4$) has a higher \ac{COG} ($2.274 \pm 1.174$) than the Euler and Midpoint configurations at the same \ac{NFE}, while also having the highest \ac{DPP} ($0.981 \pm 0.025$). Euler $s=16$ achieves \ac{COG}$=1.182 \pm 3.126$ and \ac{MMD}$=0.0283 \pm 0.0021$, compared with \ac{COG}$=1.173 \pm 3.100$ and \ac{MMD}$=0.0286 \pm 0.0021$ for Euler $s=32$. On \texttt{heatconduction2d}, increasing the number of Euler steps also does not reduce \ac{COG} monotonically. Reducing the Euler budget from 32 to 16 evaluations approximately doubles throughput, reaching $53.156 \pm 0.777$ samples/s on \texttt{beams2d} and $24.444 \pm 0.274$ samples/s on \texttt{heatconduction2d}. These values correspond to approximately $65.8\times$ and $64.4\times$ the measured Diffusion throughput, respectively. Euler $s=16$ therefore halves the measured generation time relative to $s=32$, while the reported \texttt{beams2d} \ac{COG} values differ by $0.009$.

\section{Discussion}
\label{sec:discussion}

\paragraph{Optimization utility and constraint adherence.}
Among the evaluated EngiOpt implementations, \ac{CFM} achieves the lowest measured \ac{COG} on both tasks and higher measured throughput than Diffusion. Its lower \ac{COG} indicates that the warm-started optimization trajectories accumulate a smaller objective gap relative to the paired reference designs. On \texttt{beams2d}, \ac{CFM} reaches a slightly lower final objective than Diffusion (\ac{FOG}$=-1.647$ versus $-1.637$), although the difference is small relative to the variation across seeds. On \texttt{heatconduction2d}, \ac{CFM} also achieves the lowest values for both \ac{COG} and \ac{FOG}. Its measured advantages are therefore a lower cumulative objective gap during fixed-budget optimization and lower generation time than Diffusion; their \texttt{beams2d} \ac{FOG} values differ by $0.010$.

\ac{CFM} also has the lowest mean absolute volume-fraction error on both tasks. It stays within $0.4\%$ and $1.0\%$ of the target volume fraction on \texttt{beams2d} and \texttt{heatconduction2d}, respectively, while Diffusion deviates by $3.8\%$ and $11.2\%$. A warm start far from the intended material budget may require the optimizer to restore feasibility before improving the objective. This may contribute to Diffusion's \ac{COG} on \texttt{heatconduction2d} being approximately $6.54\times$ \ac{CFM}'s ($5.81\times10^{-4}$ versus $8.88\times10^{-5}$). The Diffusion baseline uses the standard $[-1,1]$ \ac{DDPM} normalization; we did not isolate the source of its remaining volume-fraction deviations.

\ac{CFM} and Diffusion use the same conditional U-Net backbone and channel schedule. This reduces differences due to model architecture and capacity, although the two methods still differ in their training objectives and sampling procedures. Diffusion achieves higher \ac{DPP} than \ac{CFM} on both tasks, but also has higher \ac{COG}. In these experiments, higher measured diversity therefore does not correspond to better warm-start optimization trajectories. Diversity should instead be interpreted together with feasibility and optimization-utility metrics.

\paragraph{Behavior of the \acs{cGAN} baseline.}
The evaluated \ac{cGAN} is the existing EngiOpt adversarial baseline and has the highest measured sampling throughput. On \texttt{beams2d}, the large variation in \ac{COG} across seeds ($22.493 \pm 819.282$) indicates sensitivity to training initialization. On \texttt{heatconduction2d}, its \ac{COG} is $5.50\times10^{-4}$, compared with $5.81\times10^{-4}$ for Diffusion, while its \ac{DPP} is lower ($6.28\times10^{-2}\pm9.81\times10^{-2}$ versus $2.91\times10^{-1}\pm3.54\times10^{-1}$). The \ac{cGAN} therefore ranks differently by \ac{COG} and \ac{DPP} on the two tasks. These experiments do not isolate whether the observed differences arise from the conditioning variables, the datasets, or adversarial training.

\paragraph{Flow geometry and solver sensitivity.}
The solver ablation tests whether higher-order \ac{ODE} integration improves \ac{CFM} sampling at a fixed \ac{NFE}. For each noise--data pair, independent conditional flow matching uses a linear training path with constant target velocity. Because random pairings can produce crossing paths, however, the learned marginal vector field need not be straight. In the evaluated settings, Midpoint and \ac{RK4} do not improve the measured optimization metrics over Euler at matched \ac{NFE}. Increasing the Euler budget from $s=16$ to $s=32$ produces \ac{COG} values of $1.182 \pm 3.126$ and $1.173 \pm 3.100$, respectively. Euler $s=16$ achieves about $65.8\times$ the measured Diffusion throughput on \texttt{beams2d} and $64.4\times$ on \texttt{heatconduction2d}, and halves generation time relative to Euler $s=32$ in the same timing benchmark. A denser \ac{NFE} sweep would be needed to draw broader conclusions about the numerical integration methods.

\paragraph{Limitations and future work.}
The study is restricted to two 2D EngiBench tasks and one shared U-Net family for \ac{CFM} and Diffusion. The results therefore compare the evaluated EngiOpt implementations rather than establish a universal ranking of flow-matching, diffusion, and adversarial inverse-design models. The remaining Diffusion volume-fraction deviations may reflect the denoising objective, sampling dynamics, model selection, or task-specific sensitivity; isolating these factors would require targeted ablations. Future work should test whether the lower \ac{CFM} volume-fraction errors observed here persist in 3D settings, multi-physics design tasks, and higher-dimensional condition spaces.

\section{Conclusion}
\label{sec:conclusion}

This work adds \ac{CFM} to EngiOpt and evaluates it as a warm-start prior for two EngiBench inverse-design tasks. Among the tested implementations, \ac{CFM} achieves the lowest measured \ac{COG}, \ac{FOG}, \ac{MMD}, and volume-fraction deviation on both \texttt{beams2d} and \texttt{heatconduction2d}. Depending on the solver budget and task, its measured sampling throughput is approximately $32\times$--$66\times$ that of the evaluated Diffusion baseline. Reducing the Euler budget from $s=32$ to $s=16$ approximately doubles throughput, while the measured \texttt{beams2d} \ac{COG} changes from $1.173 \pm 3.100$ to $1.182 \pm 3.126$. The metric rankings are not identical: Diffusion has the highest \ac{DPP} on both tasks, whereas \ac{CFM} has the lowest \ac{COG}.

\section*{Code and Artifact Availability}

The implementation and public reproduction workflow are available in a
\href{https://github.com/IDEALLab/EngiOpt/tree/conditional-flow-matching-reproduction-v1.0.0}{tagged EngiOpt source snapshot}.
Run-level training, checkpoint-selection, evaluation, and timing records are
available in a
\href{https://wandb.ai/smassoudi-eth-z-rich/engiopt-flow-matching/reports/Conditional-Flow-Matching-for-ML-Based-Inverse-Design-Problems---Experiments-and-Reproducibility--VmlldzoxNzgzMTUyMQ==}{public Weights \& Biases report}.
The selected evaluation checkpoints and EngiBench datasets used in this study
are available through a
\href{https://huggingface.co/collections/IDEALLab/conditional-flow-matching-for-engineering-inverse-design-6a9328bf1c39d74d4460ba46}{public Hugging Face collection}.

\bibliographystyle{unsrtnat}
\bibliography{references}
\appendix
\section{Training Configuration}
\label{app:training-configuration}

All models are trained on a single NVIDIA RTX~4090 GPU with a batch size of 32 for up to 500 epochs.
\begin{itemize}
\item \textbf{\ac{CFM} and Diffusion:} Optimized with AdamW~\citep{loshchilov2017decoupled} at a learning rate of $4\times10^{-4}$, with $(\beta_1, \beta_2) = (0.9, 0.999)$ for both models.
\item \textbf{\ac{cGAN}:} Optimized with Adam~\citep{kingma2014adam} with learning rates $1\times10^{-4}$ and $4\times10^{-4}$ for the generator and discriminator, respectively, both with $(\beta_1, \beta_2) = (0.5, 0.999)$.
\end{itemize}

\section{\acs{ODE} Solver and Step-Count Ablation}
\label{app:ablation}

We ablate Euler, Midpoint, and \ac{RK4} across matched \ac{NFE} budgets of 16, 32, and 48. Because Midpoint and \ac{RK4} require two and four velocity evaluations per integration step, respectively, the corresponding step counts are adjusted to hold \ac{NFE} fixed.

\begin{table}[H]
  \centering
  \small
  \resizebox{\textwidth}{!}{\begin{tabular}{llcccc}
\toprule
Solver & Steps & \multicolumn{2}{c}{\texttt{Beams2D}} & \multicolumn{2}{c}{\texttt{HeatConduction2D}} \\
\cmidrule(lr){3-4} \cmidrule(lr){5-6}
 & & \acs{COG} $\downarrow$ & \acs{FOG} $\downarrow$ & \acs{COG} $\downarrow$ & \acs{FOG} $\downarrow$ \\
\midrule
\multirow{3}{*}{Euler}
 & 16 & $1.182 \pm 3.126$ & $-1.625 \pm 0.395$ & $1.09\times10^{-4} \pm 3.79\times10^{-5}$ & $1.34\times10^{-6} \pm 6.57\times10^{-7}$ \\
 & 32 & $1.173 \pm 3.100$ & $-1.647 \pm 0.411$ & $8.88\times10^{-5} \pm 6.84\times10^{-5}$ & $1.32\times10^{-6} \pm 8.87\times10^{-7}$ \\
 & 48 & $1.140 \pm 1.223$ & $-1.633 \pm 0.412$ & $1.70\times10^{-4} \pm 8.10\times10^{-5}$ & $1.52\times10^{-6} \pm 1.15\times10^{-6}$ \\
\midrule
\multirow{3}{*}{Midpoint}
 & 8 & $1.148 \pm 3.156$ & $-1.614 \pm 0.416$ & $1.04\times10^{-4} \pm 4.65\times10^{-5}$ & $1.49\times10^{-6} \pm 9.11\times10^{-7}$ \\
 & 16 & $1.137 \pm 2.123$ & $-1.643 \pm 0.409$ & $9.69\times10^{-5} \pm 2.94\times10^{-5}$ & $1.09\times10^{-6} \pm 7.56\times10^{-7}$ \\
 & 24 & $1.134 \pm 2.219$ & $-1.647 \pm 0.439$ & $1.14\times10^{-4} \pm 8.07\times10^{-5}$ & $2.09\times10^{-6} \pm 1.41\times10^{-6}$ \\
\midrule
\multirow{3}{*}{\acs{RK4}}
 & 4 & $2.274 \pm 1.174$ & $-1.644 \pm 0.439$ & $1.38\times10^{-4} \pm 4.64\times10^{-5}$ & $1.40\times10^{-6} \pm 7.79\times10^{-7}$ \\
 & 8 & $1.537 \pm 2.283$ & $-1.652 \pm 0.443$ & $1.13\times10^{-4} \pm 3.14\times10^{-5}$ & $1.34\times10^{-6} \pm 5.75\times10^{-7}$ \\
 & 12 & $1.263 \pm 2.320$ & $-1.599 \pm 0.415$ & $1.15\times10^{-4} \pm 5.31\times10^{-5}$ & $1.43\times10^{-6} \pm 7.67\times10^{-7}$ \\
\bottomrule
\end{tabular}
}
  \caption{Primary optimization metrics across \acs{ODE} solvers and integration budgets. \acs{COG} is reported as median $\pm$ standard deviation; \acs{FOG} is reported as mean $\pm$ standard deviation over 10 seeds. Lower is better.}
  \label{tab:ablation-cog-fog}
\end{table}

\begin{table}[H]
  \centering
  \small
  \begin{tabular}{llccc}
\toprule
Solver & Steps & \acs{MMD} $\downarrow$ & \acs{DPP} $\uparrow$ & Viol. $\downarrow$ \\
\midrule
\multicolumn{5}{l}{\textit{\texttt{Beams2D}}} \\
\midrule
\multirow{3}{*}{Euler}
 & 16 & $2.83\times10^{-2} \pm 2.13\times10^{-3}$ & $4.02\times10^{-1} \pm 2.66\times10^{-1}$ & $3.57\times10^{-3} \pm 6.22\times10^{-4}$ \\
 & 32 & $2.86\times10^{-2} \pm 2.06\times10^{-3}$ & $4.34\times10^{-1} \pm 2.69\times10^{-1}$ & $3.60\times10^{-3} \pm 6.16\times10^{-4}$ \\
 & 48 & $2.84\times10^{-2} \pm 2.10\times10^{-3}$ & $4.33\times10^{-1} \pm 2.73\times10^{-1}$ & $3.33\times10^{-3} \pm 5.57\times10^{-4}$ \\
\midrule
\multirow{3}{*}{Midpoint}
 & 8 & $2.87\times10^{-2} \pm 1.96\times10^{-3}$ & $4.65\times10^{-1} \pm 2.66\times10^{-1}$ & $3.55\times10^{-3} \pm 4.48\times10^{-4}$ \\
 & 16 & $2.89\times10^{-2} \pm 1.86\times10^{-3}$ & $4.76\times10^{-1} \pm 2.68\times10^{-1}$ & $3.67\times10^{-3} \pm 5.63\times10^{-4}$ \\
 & 24 & $2.97\times10^{-2} \pm 1.68\times10^{-3}$ & $4.72\times10^{-1} \pm 2.74\times10^{-1}$ & $3.86\times10^{-3} \pm 6.67\times10^{-4}$ \\
\midrule
\multirow{3}{*}{\acs{RK4}}
 & 4 & $3.96\times10^{-2} \pm 1.63\times10^{-3}$ & $9.81\times10^{-1} \pm 2.52\times10^{-2}$ & $9.85\times10^{-3} \pm 2.54\times10^{-3}$ \\
 & 8 & $3.50\times10^{-2} \pm 1.57\times10^{-3}$ & $8.83\times10^{-1} \pm 1.44\times10^{-1}$ & $5.33\times10^{-3} \pm 4.81\times10^{-4}$ \\
 & 12 & $3.20\times10^{-2} \pm 1.64\times10^{-3}$ & $7.82\times10^{-1} \pm 1.81\times10^{-1}$ & $4.62\times10^{-3} \pm 4.55\times10^{-4}$ \\
\midrule
\multicolumn{5}{l}{\textit{\texttt{HeatConduction2D}}} \\
\midrule
\multirow{3}{*}{Euler}
 & 16 & $5.80\times10^{-2} \pm 8.79\times10^{-3}$ & $7.35\times10^{-3} \pm 1.65\times10^{-2}$ & $1.03\times10^{-2} \pm 2.08\times10^{-3}$ \\
 & 32 & $5.75\times10^{-2} \pm 8.35\times10^{-3}$ & $8.99\times10^{-3} \pm 1.35\times10^{-2}$ & $9.96\times10^{-3} \pm 2.09\times10^{-3}$ \\
 & 48 & $5.89\times10^{-2} \pm 7.57\times10^{-3}$ & $4.49\times10^{-2} \pm 8.66\times10^{-2}$ & $9.79\times10^{-3} \pm 1.79\times10^{-3}$ \\
\midrule
\multirow{3}{*}{Midpoint}
 & 8 & $5.75\times10^{-2} \pm 7.83\times10^{-3}$ & $2.38\times10^{-3} \pm 2.63\times10^{-3}$ & $1.01\times10^{-2} \pm 1.85\times10^{-3}$ \\
 & 16 & $5.61\times10^{-2} \pm 8.63\times10^{-3}$ & $7.37\times10^{-3} \pm 1.26\times10^{-2}$ & $1.01\times10^{-2} \pm 1.59\times10^{-3}$ \\
 & 24 & $5.87\times10^{-2} \pm 1.03\times10^{-2}$ & $2.45\times10^{-2} \pm 5.00\times10^{-2}$ & $1.00\times10^{-2} \pm 1.54\times10^{-3}$ \\
\midrule
\multirow{3}{*}{\acs{RK4}}
 & 4 & $5.76\times10^{-2} \pm 9.58\times10^{-3}$ & $2.65\times10^{-2} \pm 3.74\times10^{-2}$ & $9.83\times10^{-3} \pm 2.57\times10^{-3}$ \\
 & 8 & $5.70\times10^{-2} \pm 7.63\times10^{-3}$ & $1.79\times10^{-2} \pm 3.27\times10^{-2}$ & $9.61\times10^{-3} \pm 1.89\times10^{-3}$ \\
 & 12 & $5.91\times10^{-2} \pm 1.01\times10^{-2}$ & $3.46\times10^{-2} \pm 7.03\times10^{-2}$ & $1.01\times10^{-2} \pm 2.36\times10^{-3}$ \\
\bottomrule
\end{tabular}

  \caption{Distributional and feasibility metrics for the solver-step ablation (mean $\pm$ standard deviation over 10 seeds). Lower is better for \acs{MMD} and Viol.; higher is better for \acs{DPP}.}
  \label{tab:ablation-secondary}
\end{table}

\begin{table}[H]
  \centering
  \small
  \begin{tabular}{llcc}
\toprule
Solver & Problem & gen\_runtime\_sec $\downarrow$ & samples/s $\uparrow$ \\
\midrule
\multirow{2}{*}{Euler $s=16$}
  & \texttt{beams2d} & $0.941 \pm 0.014$ & $53.156 \pm 0.777$ \\
  & \texttt{heatconduction2d} & $2.046 \pm 0.023$ & $24.444 \pm 0.274$ \\
\midrule
\multirow{2}{*}{Euler $s=32$}
  & \texttt{beams2d} & $1.876 \pm 0.022$ & $26.659 \pm 0.312$ \\
  & \texttt{heatconduction2d} & $4.092 \pm 0.052$ & $12.221 \pm 0.157$ \\
\midrule
\multirow{2}{*}{Euler $s=48$}
  & \texttt{beams2d} & $2.813 \pm 0.025$ & $17.774 \pm 0.157$ \\
  & \texttt{heatconduction2d} & $6.128 \pm 0.057$ & $8.160 \pm 0.076$ \\
\midrule
\multirow{2}{*}{Midpoint $s=8$}
  & \texttt{beams2d} & $0.938 \pm 0.011$ & $53.293 \pm 0.634$ \\
  & \texttt{heatconduction2d} & $2.032 \pm 0.017$ & $24.602 \pm 0.208$ \\
\midrule
\multirow{2}{*}{Midpoint $s=16$}
  & \texttt{beams2d} & $1.880 \pm 0.021$ & $26.592 \pm 0.297$ \\
  & \texttt{heatconduction2d} & $4.062 \pm 0.034$ & $12.311 \pm 0.103$ \\
\midrule
\multirow{2}{*}{Midpoint $s=24$}
  & \texttt{beams2d} & $2.798 \pm 0.022$ & $17.874 \pm 0.141$ \\
  & \texttt{heatconduction2d} & $6.106 \pm 0.058$ & $8.189 \pm 0.078$ \\
\midrule
\multirow{2}{*}{\acs{RK4} $s=4$}
  & \texttt{beams2d} & $0.942 \pm 0.011$ & $53.102 \pm 0.633$ \\
  & \texttt{heatconduction2d} & $2.030 \pm 0.023$ & $24.636 \pm 0.275$ \\
\midrule
\multirow{2}{*}{\acs{RK4} $s=8$}
  & \texttt{beams2d} & $1.882 \pm 0.025$ & $26.572 \pm 0.357$ \\
  & \texttt{heatconduction2d} & $4.060 \pm 0.038$ & $12.317 \pm 0.115$ \\
\midrule
\multirow{2}{*}{\acs{RK4} $s=12$}
  & \texttt{beams2d} & $2.809 \pm 0.023$ & $17.801 \pm 0.148$ \\
  & \texttt{heatconduction2d} & $6.111 \pm 0.056$ & $8.183 \pm 0.075$ \\
\bottomrule
\end{tabular}

  \caption{Sampling throughput across \acs{ODE} solvers and integration steps (mean $\pm$ standard deviation over 10 seeds).}
  \label{tab:ablation-efficiency}
\end{table}

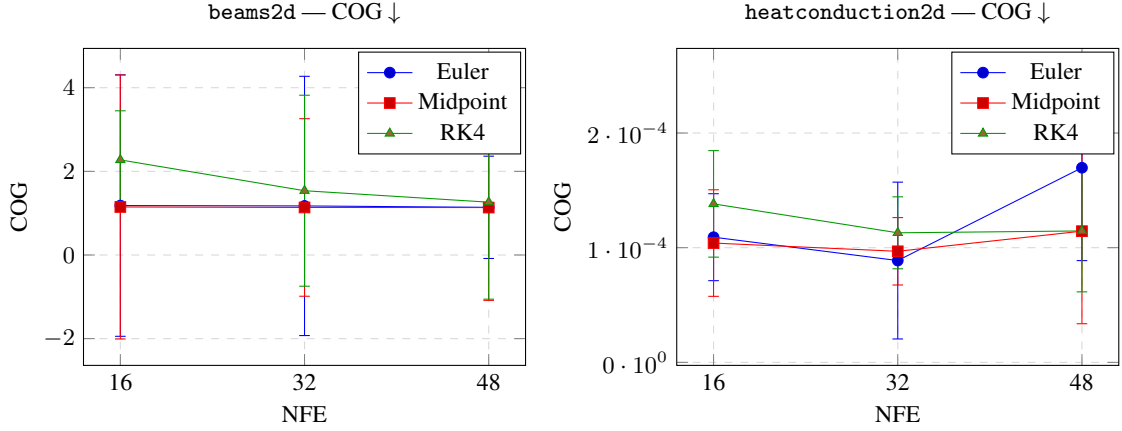
\begin{figure}[htbp]
  \centering
  \small
  \begin{tikzpicture}
\begin{groupplot}[
    group style={group size=2 by 1, horizontal sep=2cm},
    width=0.45\textwidth,
    height=0.35\textwidth,
    xlabel={\acs{NFE}},
    xtick={16,32,48},
    xticklabels={16, 32, 48},
    legend style={at={(0.99,0.99)}, anchor=north east, font=\small},
    grid=major,
    grid style={dashed, gray!30},
]
\nextgroupplot[title={\texttt{beams2d} --- \acs{COG} $\downarrow$}, ylabel={\acs{COG}}]
\addplot+[mark=*, color=blue, error bars/.cd, y dir=both, y explicit]
    coordinates {(16,1.18170537)+-(0,3.1261456) (32,1.17320369)+-(0,3.09951706) (48,1.13961479)+-(0,1.22316941)};
\addlegendentry{Euler}
\addplot+[mark=square*, color=red, error bars/.cd, y dir=both, y explicit]
    coordinates {(16,1.14835166)+-(0,3.15644006) (32,1.13676212)+-(0,2.12285876) (48,1.13428982)+-(0,2.21927656)};
\addlegendentry{Midpoint}
\addplot+[mark=triangle*, color=green!60!black, error bars/.cd, y dir=both, y explicit]
    coordinates {(16,2.27417327)+-(0,1.17383653) (32,1.53697679)+-(0,2.28324542) (48,1.26327347)+-(0,2.31999499)};
\addlegendentry{\acs{RK4}}
\nextgroupplot[title={\texttt{heatconduction2d} --- \acs{COG} $\downarrow$}, ylabel={\acs{COG}},
    scaled y ticks=false,
    yticklabel={\pgfmathprintnumber[sci, precision=1]{\tick}}]
\addplot+[mark=*, color=blue, error bars/.cd, y dir=both, y explicit]
    coordinates {(16,0.00010914964)+-(0,3.79077043e-05) (32,8.87751585e-05)+-(0,6.83872804e-05) (48,0.000169734805)+-(0,8.10043704e-05)};
\addlegendentry{Euler}
\addplot+[mark=square*, color=red, error bars/.cd, y dir=both, y explicit]
    coordinates {(16,0.000104034731)+-(0,4.64572562e-05) (32,9.68542081e-05)+-(0,2.93898657e-05) (48,0.000114410675)+-(0,8.07312532e-05)};
\addlegendentry{Midpoint}
\addplot+[mark=triangle*, color=green!60!black, error bars/.cd, y dir=both, y explicit]
    coordinates {(16,0.000138202511)+-(0,4.64458724e-05) (32,0.000112977643)+-(0,3.13765112e-05) (48,0.000114603483)+-(0,5.311505e-05)};
\addlegendentry{\acs{RK4}}
\end{groupplot}
\end{tikzpicture}
  \caption{Ablation of \acs{ODE} solver families across \acs{NFE} budgets. At low \acs{NFE}, \acs{RK4} uses fewer, larger integration steps and produces higher \acs{COG} than Euler on \texttt{beams2d}.}
  \label{fig:ablation-solver}
\end{figure}

On \texttt{beams2d}, Euler and Midpoint produce \ac{COG} values between $1.134$ and $1.182$ across all \ac{NFE} settings. \ac{RK4} at \ac{NFE}$=16$ ($s=4$) has the highest \ac{COG} ($2.274 \pm 1.174$), \ac{DPP} ($0.981 \pm 0.025$), and \ac{MMD} ($0.040 \pm 0.002$) among the matched-\ac{NFE} settings. Its greater measured diversity therefore does not coincide with a lower warm-start \ac{COG}. At \ac{NFE}$=32$, \ac{RK4} ($s=8$) has \ac{COG}$=1.537$, compared with $1.173$ for Euler $s=32$. On \texttt{heatconduction2d}, \ac{COG} varies between $8.88\times10^{-5}$ and $1.70\times10^{-4}$ across the evaluated configurations.

The \ac{NFE}--\ac{COG} relationship is non-monotonic for Euler on \texttt{beams2d}. Euler $s=16$ has \ac{COG}$=1.182 \pm 3.126$ and \ac{MMD}$=0.0283 \pm 0.0021$, compared with \ac{COG}$=1.173 \pm 3.100$ and \ac{MMD}$=0.0286 \pm 0.0021$ for Euler $s=32$, while its throughput increases to $53.156$ samples/s.

\section*{Acronyms}
\begin{acronym}[I-CFM]
  \acro{CFM}{conditional flow matching}
  \acro{COG}{cumulative optimality gap}
  \acro{FOG}{final optimality gap}
  \acro{NFE}{number of function evaluations}
  \acro{cGAN}{conditional generative adversarial network}
  \acro{ICFM}[I-CFM]{independent conditional flow matching}
  \acro{GAN}{generative adversarial network}
  \acro{MMD}{maximum mean discrepancy}
  \acro{DPP}{determinantal point process}
  \acro{ODE}{ordinary differential equation}
  \acro{RK4}{classical fourth-order Runge--Kutta method}
  \acro{DDPM}{denoising diffusion probabilistic model}
  \acro{PDE}{partial differential equation}
\end{acronym}

\end{document}